# Appropriate and Inappropriate Estimation Techniques

David Sher
Computer Science Dept, University of Rochester, Rochester NY 14627

## ABSTRACT

Mode (also called MAP) estimation, mean estimation and median estimation are examined here to determine when they can be safely used to derive (posterior) cost minimizing estimates. (These are all Bayes procedures, using the mode, mean, or median of the posterior distribution.) It is found that modal estimation only returns cost minimizing estimates when the cost function is 0-1. If the cost function is a function of distance then mean estimation only returns cost minimizing estimates when the cost function is squared distance from the true value and median estimation only returns cost minimizing estimates when the cost function is the distance from the true value. Results are presented on the goodness of modal estimation with non 0-1 cost functions.

In low-level vision small-scale image phenomena are often assigned labels such as "on a boundary" or "moving at .3 pixels per frame". Most current approaches to low-level vision return estimates of the feature labeling [Ballard82] [Andrews77b]. Any such technique operates implicitly by deriving an estimate from a posterior distribution. Given that the correct posterior distribution is derived an estimation technique, such as returning the label with highest probability, returns the most useful estimates when the user has an appropriate cost function. A cost function is *appropriate* for an estimation technique if the estimation technique returns the expected cost minimizing estimate. The question addressed in this paper is: "What are the appropriate cost functions for different estimation techniques and how much extra cost is incurred by using an estimation technique with an inappropriate cost function?" This work is motivated by and lies within the field of decision theory [Berger80b].

## 1. Previous Work and Estimation

Some algorithms for low-level vision return the feature labeling with the highest probability. This estimation technique is commonly called *modal estimation*. Work that has been done based on modal estimation is simulated annealing work [Geman84], image reconstruction work [Andrews77c] [Jaynes85] [Herman85], and hough transform work [Hough62] [Ballard81] [Brown83] [Turney85].

Returning the expected label is an estimation technique for parametric label spaces. This estimation technique is commonly called *mean estimation*. Least squares techniques use mean estimation [Andrews77a] [Herman85] [Turney85]. One of the best known approaches that uses least squares and mean estimation is the Wiener filter [Andrews77a].

Another possible approach to estimation is to return the median of the distribution. Such an approach is called *median estimation*. The only vision algorithm I know of that uses the median is median filtering [Fisher81].

A work comparing a technique for modal estimation with a technique for mean estimation has been written by Marroquin, Mitter and Poggio [Marroquin85]. They observe that modal estimation can be less appropriate for image reconstruction problems than mean estimation.

Much work in statistics is on estimation problems. The usual problem approached is to derive given a known cost function the optimal estimation procedure. Thus it is well known that the mean is optimal for a squared distance cost function, the median for a absolute difference cost function and the mode for a 0-1 cost function [Berger80c].

## 2. Terminology

An estimation problem has a set of states of the universe $S$ and observed data $O$. An *estimator* reports some state $s \in S$ from the data $O$. The universe is really in some state $t \in S$ that may be in fact be $s$. This state, $t$, is the *state of nature*. Given the observed data $O$ a probability distribution can be derived over $S$. This probability distribution I call the *posterior distribution* over $S$ and refer to $P(s)$ (ignoring $O$ from this point further). There is a cost for an estimator reporting state $s$ when state $t$ is the state of nature. This cost is described by a *cost function* $\delta(s,t)$.

The purpose of an estimation problem is to derive the true state of nature and to minimize the cost of a mistake. Thus for any $t$ equation 1 must hold.

261

$$\delta(t,t) = \min_{s \in S} \delta(s,t) \qquad (1)$$

This paper examines the structure of *expected posterior costs of estimates*. The *expected posterior cost* of an estimate is the cost of the estimate (based on the data) integrated over possible states of nature (equation 2).

$$E(\delta(s,t)) = \int_{t \in S} \delta(s,t) dP(t) \qquad (2)$$

Thus if two different cost functions $\delta_1$ and $\delta_2$ result in the same inequalities between expected costs of actions they can be considered *equivalent cost functions*. The *expected prior cost of the estimation technique*, the expected cost before data has been collected, is minimized when the expected posterior cost is minimized.

Any increasing linear transformation of a cost function that results in an equivalent cost function. In estimation problems any cost function $\delta$ has an equivalent cost $\delta'$ such that $\forall t\, \delta'(t,t) = 0$ [Berger80a]. In this paper I only consider everywhere continuous cost functions over compact sets. Hence I can safely say that for every cost function $\delta$ there is an equivalent one $\delta'$ such that $\forall t\, \delta'(t,t) = 0$ and $\max_{s,t} \delta(s,t) = 1$. Thus I need only consider cost functions of this form.

### 3. Relative Error

A technique for measuring extra costs is necessary. The measure should be the same for equivalent cost functions. A measure that almost has this property is the maximal relative error of costs. Assume that an estimation procedure derives estimate $e(P)$ for posterior distribution $P$ and the expected cost minimizing estimate is $s$. Then the *relative error subject to $P$* is in equation 3.

$$\frac{E(\delta(e(P),t)) - E(\delta(s,t))}{E(\delta(s,t))} \qquad (3)$$

The relative error is invariant under multiplication by a constant of the cost function. It is assumed when deriving the relative error that $\forall t\, \delta(t,t) = 0$.

In this paper I examine the maximum over all $P$ of the relative error subject to $P$. I call this maximum *the relative error of the estimation technique*. If it is 0 then the cost function is appropriate for the estimation procedure.

### 4. Mode estimation

In this paper I examine modal estimation (also often called *MAP estimation*) to find what cost functions are appropriate and to determine what happens when the cost function fails to be appropriate.

#### 4.1. Symmetry of Costs

If MAP is an appropriate estimation technique $\delta$ must be symmetric (equation 4).

$$\delta(s,t) = \delta(t,s) \qquad (4)$$

Given a probability distribution that concentrates all the probability of the distribution on the two values $s$, and $t$, and we let $P(t)$ approach $P(s)$ from below $s$ continues to be the mode of the distribution and equation continues to hold. Here equation 4 approaches in the limit equation 5.

$$\frac{\delta(s,t)}{\delta(t,s)} - 1 \qquad (5)$$

Thus equation 5 is a lower limit on the relative error of mode estimation. Thus if modal estimation is appropriate the cost function is symmetric.

As an example consider a game of coin guessing with unfair coins where guessing correctly wins $1 and guessing heads when tails looses $2 and guessing tails when heads loses $1. Then with modal estimation there is as much as 50% more cost from playing the game than with the minimum cost estimate. (Since the difference from a correct guess is $-3 for heads when tails but $-2 for tails when heads giving a relative error of .5.)

#### 4.2. Equivalence

In this section I develop another restriction on $\delta$ for mode to be appropriate (in equation 6).

$$\delta(s,u) = 0 \Rightarrow \forall t\, \delta(s,t) = \delta(u,t) \qquad (6)$$

Consider a $P$ that concentrates all the probability on $s, t,$ and $u$. Let $s$ have the highest probability. Let $t$ have positive but arbitrarily small probability. If $\delta(s,t) > \delta(u,t)$ then $u$ is the cost minimizing estimate. The relative error is in equation 7.

262

$$\frac{\delta(s,t)P(t)}{\delta(u,t)P(t)} - 1 = \frac{\delta(s,t)}{\delta(u,t)} - 1$$

It is more difficult to find a convincing example of a situation where equation 6 is violated. Consider flipping 2 coins where the flips are possibly dependent and the coins are not necessarily fair. The attempt is to estimate what the flip was and the cost for mistakes is as in figure 1.

| State | HH | HT | TH | TT |
|-------|----|----|----|----|
| HH    | 0  | 0  | 1  | 1  |
| HT    | 0  | 0  | 2  | 1  |
| TH    | 1  | 2  | 0  | 1  |
| TT    | 1  | 1  | 1  | 0  |

Figure 1: Cost Function for Two Coin Game

Guessing TH instead of HH costs $1. But guessing TH instead of HT costs $2. Thus modal estimation can result in choosing HT when HH is the best choice because HT is slightly more probable than HH.

Using equation 6 and symmetry for $\delta$ I can show that $\delta(s,t) = 0$ is an equivalence relationship on $S$ when modal estimation is appropriate. Also it is easy to show that appropriate $\delta$ respect equivalence classes in $S$.

### 4.3. All Positive Costs are Unity

When modal estimation is appropriate all positive values of an appropriate $\delta$ are shown to be equal to a constant. Since all increasing linear functions of $\delta$ are equivalent to $\delta$ I can assume without loss of generality that all positive values of $\delta$ equal 1.

To derive the positive lower limit on the relative error a $P$ is assumed to place all its probability on 3 elements of $S$, $s,t$, and $u$. To derive the lower limit assume that $\delta(s,t)>0$ and $\delta(t,u)>0$ and without loss of generality $\delta(s,t)<\delta(t,u)$.

Consider what happens when the probabilities of $s,t$, and $u$ all approach each other (get within $\varepsilon$ of each other). If $\delta(s,u)<\delta(t,u)$, $s$ is the cost minimizing estimate. If the probabilities of $s$ and $t$ approach the probability $u$ from below the relative error in the modal estimate is given in equation 8.

$$\frac{\delta(s,u)+\delta(u,t)}{\delta(s,u)+\delta(s,t)} - 1 \qquad (8)$$

If $\delta(s,u) \geq \delta(t,u)$, $t$ is the cost minimizing estimate. Thus if the probabilities of $s$ and $t$ approach the probability of $u$ from below here equation 9 describes the relative error in the modal estimation.

$$\frac{\delta(s,u)+\delta(u,t)}{\delta(s,t)+\delta(u,t)} - 1 \qquad (9)$$

The numbers in equations 8 and 9 are positive. Thus equation 8 and 9 describe a lower limit on the relative error for $\delta$ when there are two positive values of $\delta$ that are not equal. Hence all positive values of $\delta$ must be equal when it is appropriate. Thus without loss of generality I can assume that all positive costs are 1.

As an example consider an estimation problem where the possible states are 0, 1, and 2. The cost of an estimate is the absolute difference of the estimate and the state of nature. Let $s$ correspond to 1, $t$ correspond to 0 and $u$ correspond to 2. $\delta(s,t) = 1$, $\delta(t,u) = 2$ and $\delta(s,u)<\delta(t,u)$. Hence the relative error in this problem is at least $\frac{1}{2}$ from equation 8. This error occurs when $P(0)+\varepsilon = P(1)+\varepsilon = P(2)$.

### 4.4. If Two Events Have Cost Zero They All Do

It is simple to show that if two elements $s,t$ of $S$ have $\delta(s,t)=0$ and two elements $u$, $v$ have $\delta(u,v) = 1$ then the cost function is not appropriate for modal estimation. Assume $\delta$ conforms to the necessary conditions for modal estimation established in the previous sections. Then every value of $\delta$ is 0 or 1. Thus if $s,t \in S$ have $\delta(s,t) = 0$ then there are two elements of $S$, $s$ and $t$, such that there is an element of $S$, $u$, such that $\delta(s,t) = 0$ and $\delta(s,u) = \delta(t,u) = 1$ (by the pigeon hole principle).

Consider a $P$ that concentrates all the probability onto $s,t$, and $u$. If $u$ is the mode and $s$ (or $t$) minimizes the costs the relative error has a lower limit written in equation 10.

$$\frac{P(s)+P(t)}{P(u)} - 1 \qquad (10)$$

If the probabilities of $s$ and $t$ approach the probability of $u$ from below the conditions for equation 10 are satisfied and the relative error described by it goes to 1.

263

When modal estimation is appropriate then there are two possible cost functions the all 0 function and the function that is 1 except when the estimate is correct. The all 0 function is useless so the only useful cost function that has modal estimation appropriate is the one that is 1 except when the estimate is correct. This function is often referred to as a 0 1 cost function.

Open questions are what happens when several different errors appear in the same cost function and whether these errors can compound each other to get relative errors greater than those predicted here.

## 5. Mean Estimation

To analyze mean estimation I need $S$ to be a Banach space. I assumed in section 2 that $S$ is compact. To make the problem mathematically tractable assume that $S$ is a closed ball in $\mathbf{R}^n$. Also assume that $\delta$ is differentiable and bounded everywhere.

Consider a $P$ that places all its probability on a line, $L$. Thus I can characterize the problem as equation 11.

$$\frac{d}{de} \int_{S \cap L} \delta(e,t) dP(t) = 0 \qquad (11)$$

Also let $\delta(s,t)$ be a function of the absolute difference of $s$ and $t$ (along the line). Thus $\delta(e,t) = f(|e-t|)$ From this derives equation 12.

$$\int_{\min(S \cap L)}^{e} f'(|e-t|) dP(t) = \int_{e}^{\max(S \cap L)} f'(|e-t|) dP(t) \quad (12)$$

For what functions $f$ does the mean work as $e$ in equation 12. Without loss of generality we can assume $L$ is the real line. Let $P$ put probability $\frac{n}{n+1}$ at $-x$ and probability $\frac{1}{n+1}$ at $nx$. Equation 12 reduces to equation 13.

$$\forall x \; nf'(x) = f'(nx) \qquad (13)$$

Given this functional equation one can show that the only $\delta(s,t)$, that are appropriate for mean estimation, are ones where $\delta(s,t)$ $\delta(s,t) = L(s-t)L(s-t)^T$ for a linear function $\mathbf{L}$.

An open question is whether there are $\delta$ appropriate for mean estimation with $\delta$ not a function of the absolute difference of coordinates in some coordinate system. Most proposed cost functions have this form. An open research topic is the relative error when using mean estimation.

## 6. Median Estimation

The median can only be defined on totally ordered sets. To define the median $S$ also has to be a Banach space (so that integration is defined). $m$ is the median if equation 14 holds.

$$\int_{\min S}^{m} dP = \int_{m}^{\max S} dP \qquad (14)$$

To attain mathematical results assume that $S$ is a subset of $\mathbf{R}$. Thus I can use results derived in section 5 (where I made a weaker assumption) about the properties that an estimate must have to be cost minimizing, in particular equation 12.

$$\int_{\min(S \cap L)}^{e} f'(|e-t|) dP(t) = \int_{e}^{\max(S \cap L)} f'(|e-t|) dP(t) \quad 12$$

Clearly $f'$ must be constant to satisfy this equation for all $P$.

An open question is whether there are $\delta$ appropriate for median estimation with $\delta$ not a function of the absolute difference of coordinates in some coordinate system. Most proposed cost functions have this form. An open research topic is the relative error when using median estimation.

## 7. Conclusions

In this paper I have discussed three different forms of estimation: modal estimation, mean estimation, and median estimation. I have found necessary and sufficient conditions for modal estimation to minimize posterior costs: The cost function is either equivalent to a 0-1 cost function or trivial. If the cost function does not fit this form then modal estimation results in non-optimal estimates. Just how much extra cost can be incurred by modal estimation was discussed. It was shown that among cost functions that are functions of the distance only the least squares cost can use mean estimation for cost minimizing estimation and only absolute difference cost can use median estimation for cost minimizing estimation.

These results are useful for evaluating algorithms that use estimation. If an algorithm uses an estimation technique that is not appropriate for the application's cost function then that algorithm must be viewed with distrust as noted by Marroquin [Marroquin85]. A more expanded version of this paper will appear in an upcoming technical report.

## 8. Acknowledgements

I was greatly helped in this work by the advice of Chris Brown, Jack Hall and Lawrence Sher. This work was supported by DARPA/US Army Engineering Topographic Labs grant number DACA76-85-C-0001.

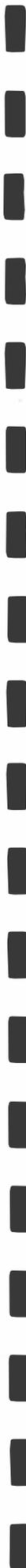